\documentclass[runningheads]{llncs}
\usepackage{graphicx}
\usepackage{amsmath}

\usepackage{amsfonts}

\usepackage{hyperref}
\hypersetup{hidelinks,
	colorlinks=true,
	allcolors=black,
	pdfstartview=Fit,
	breaklinks=true}

\usepackage{array}
\newcolumntype{M}[1]{>{\centering\arraybackslash}m{#1}}
\begin{document}
\title{Document Dewarping with Control Points}
%
%
\author{Guo-Wang Xie\inst{1,2} \and
Fei Yin\inst{2} \and
Xu-Yao Zhang\inst{1,2} \and
Cheng-Lin Liu\inst{1,2,3}}
\authorrunning{Xie et al.}
%
\institute{
School of Artificial Intelligence, University of Chinese Academy of Sciences, Beijing 100049, P.R. China\and
National Laboratory of Pattern Recognition, Institute of Automation of Chinese Academy of Sciences, 95 Zhongguancun East Road, Beijing 100190, P.R. China \and
CAS Center for Excellence of Brain Science and Intelligence Technology, Beijing, P.R. China\\
\email{xieguowang2018@ia.ac.cn}\\
\email{\{fyin, xyz, liucl\}@nlpr.ia.ac.cn}}
\maketitle              
\begin{abstract}
Document images are now widely captured by handheld devices such as mobile phones. The OCR performance on these images are largely affected due to geometric distortion of the document paper, diverse camera positions and complex backgrounds. In this paper, we propose a simple yet effective approach to rectify distorted document image by estimating control points and reference points. After that, we use interpolation method between control points and reference points to convert sparse mappings to backward mapping, and remap the original distorted document image to the rectified image. Furthermore, control points are controllable to facilitate interaction or subsequent adjustment. We can flexibly select post-processing methods and the number of vertices according to different application scenarios. Experiments show that our approach can rectify document images with various distortion types, and yield state-of-the-art performance on real-world dataset. This paper also provides a training dataset based on control points for document dewarping. Both the code and the dataset are released at \href{https://github.com/gwxie/Document-Dewarping-with-Control-Points}{https://github.com/gwxie/Document-Dewarping-with-Control-Points}.

\end{abstract}
\section{Introduction}
Document image has become very common and important in our daily life because of their convenience in archiving, retrieving and sharing valuable information. Unlike the controllable operating environment of the scanner, camera-captured document images often suffer from distortions and background, due to  physical deformation of the paper, shooting environment and camera positions. The above factors will significantly increase the difficulty of information extraction and content analysis. For reducing the influence of distortion in document image processing, many dewarping approaches have been proposed in the literature.

\begin{figure}
\includegraphics[width=\textwidth]{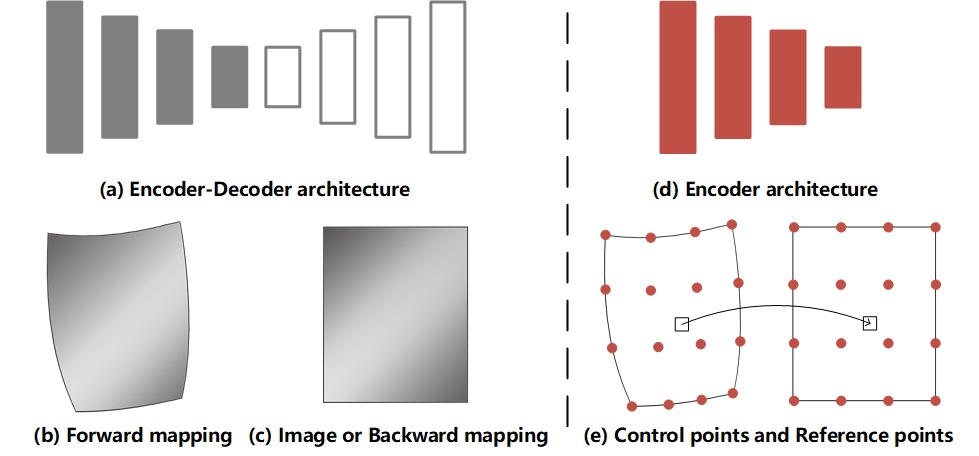}
\caption{{\bfseries Dewarping architecture.} (a) the Encoder-Decoder architecture is used as a generic feature extractor to predict the expression of the pixel-wise such as (b) forward mapping and (c) image or backward mapping in recent rectifying systems. Our proposed sparse points is based on rectification which only exploits (d) Encoder architecture to predict (e) control points and reference points so as to achieves similar effects as (a) but in a more flexible and practical way.} \label{fig1}
\end{figure}

Traditional approaches~\cite{ref_2,ref_4,ref_8} usually require complex process, external conditions or strong assumptions to construct 2D or 3D rectification models by extracting the hand-crafted features of document images. To improve generalization ability in difficult scenarios, some deep learning based approaches have been proposed and promising performance in rectification is obtained. As shown in Fig.~\ref{fig1}(a), a widely-used approach is to exploit the Encoder-Decoder architecture as a generic feature extractor to predict some pixel-wise information, such as forward mapping (each cell represents the coordinates of the pixels in the dewarped ouput image, and the pixels correspond to pixels in the warped input image)~\cite{ref_11,ref_14,ref_16} in Fig.~\ref{fig1}(b) and dewarping image (image-image translation)~\cite{ref_10} or backward mapping (each cell represents the coordinates of the pixels in the warped input image)~\cite{ref_12,ref_15,ref_13} in Fig.~\ref{fig1}(c). Although the Encoder-Decoder architecture has achieved satisfying performance, further research is needed for more flexible and lightweight approaches.

This paper proposes a novel approach to rectify distorted document image and remove background finely. As shown in Fig.~\ref{fig1}(d), we take advantage of the Encoder architecture for extracting semantic information from image automatically, which is used to predict control points and reference points in Fig.~\ref{fig1}(e). The control points and reference points are composed of the same number of vertices and describe the shape of the document in the image before and after rectifying, respectively. Then, we use interpolation method between control points and reference points to convert sparse mappings to backward mapping, and remap the original distorted document image to the rectified image. The control points are flexible and controllable, which facilitates the interaction with people to adjust the sub-optimal points. Experiments show that the method based on control points can rectify various deformed document images, and yield state-of-the-art performance on real-world dataset. Furthermore, our approach can be edited multiple times to improves its practicability when the correction effect is not satisfied, which alleviates the disadvantages of weak operability in end-to-end methods. We can flexibly select post-processing methods and the number of vertices according to different application scenarios. Compared to the pixel-wise regression, control points is more practical and efficient. To inspire future researches on this direction, we also provide a new dataset based on control points for document dewarping.

\section{Related Works}
In recent years, researchers have investigated a variety of approaches to rectify distorted document. We give a brief overview of these methods from the perspectives of handcrafted features and learning-based features.

{\bfseries Handcrafted features-based rectification.}
Prior to the prevalence of deep learning, most approaches constructed 2D and 3D rectification models by extracting hand-crafted features of document images. Some of these approaches utilized visual cues of the document image to reconstruct the document surface, such as text lines~\cite{ref_1,ref_2,ref_3}, illumination/shading~\cite{ref_4,ref_5,ref_6} etc.  Stamatopoulos et al.~\cite{ref_1} rectified the document image in a coarse scale by detecting words and text lines, and used baseline correction to further refine and normalize individual words. Tian et al.~\cite{ref_2} estimated the 2D distortion grid by identifying and tracing text lines, and then estimated the 3D deformation from the 2D distortion grid. Liu et al.~\cite{ref_3} estimated baselines’ shape and characters’ slant angles, and then exploited thinplate splines to recover the distorted image to be flat. Wada et al.~\cite{ref_4} and Courteille et al.~\cite{ref_5} employed the technique of shape from shading (SfS) and restored the distorted image based on the reconstructed 3D shape. Similarly, Zhang et al.~\cite{ref_6} proposed a generic SfS method considering the perspective projection model and various lighting conditions, and then mapped the 3D surface back to a plane. Moreover, there were many approaches that use geometric properties or  multi-view to rectify distorted document. Brown et al.~\cite{ref_8} used the 2-D boundary of the imaged material to correct geometric distortion. He et al.~\cite{ref_7} extracted page boundary curves to reconstruct the 3D surface.  Tsoi et al.~\cite{ref_9} utilized the boundary information from multiple views of the document image to recover geometric distortions. For the existence of simple skew, binder curl, and folded deformation of the document, handcrafted features-based rectification demonstrated good performance. However, these methods are difficult to be employed in dealing with distorted documents captured from natural scenes due to their complicated geometric distortion and  changeable external conditions. 

{\bfseries Learned features-based rectification.}
With the progress of deep learning research, a lot of works exploited the features learned from document image to recover geometric distortions. Ramanna et al.~\cite{ref_10} synthesized dewarped image by the deep learning network of image-image translation (pix2pixhd). Ma et al.~\cite{ref_11} created a large-scale synthetic dataset by warping non-distorted document images and proposed a stacked U-Net to predict the forward mapping for the warping. Liu et al.~\cite{ref_14} adopted adversarial network to predict a dense unwarping grid at multiple resolutions in a coarse-to-fine fashion. In order to improve generalization in real-world images, many approaches~\cite{ref_12,ref_13,ref_15} focused on generating more realistic training dataset which has a more similar distribution to the real-world image. Das and Ma et al.~\cite{ref_12} and Markovitz et al.~\cite{ref_15} exploited multiple ground-truth annotations in both 2D and 3D domain to predict the backward mapping of a warped document image. Li et al.~\cite{ref_13} proposed patch-based learning approach and stitch the patch results into the rectified document. Although higher-quality training dataset and richer ground-truth annotations make it easier for the model to learn useful features, it also increases the difficulty for engineers to build the datasets and also the models. In order to better tradeoff between model complexity and rectification performance, Xie et al.~\cite{ref_16} proposed a novel framework to estimate pixel-wise displacements and foreground/background classification. Compared to prior approaches,~\cite{ref_16} achieved better performance with various distorted document images, but there is still room for improvement in the computational complexity and post-processing steps.

\section{Approach}
\begin{figure}
\includegraphics[width=\textwidth]{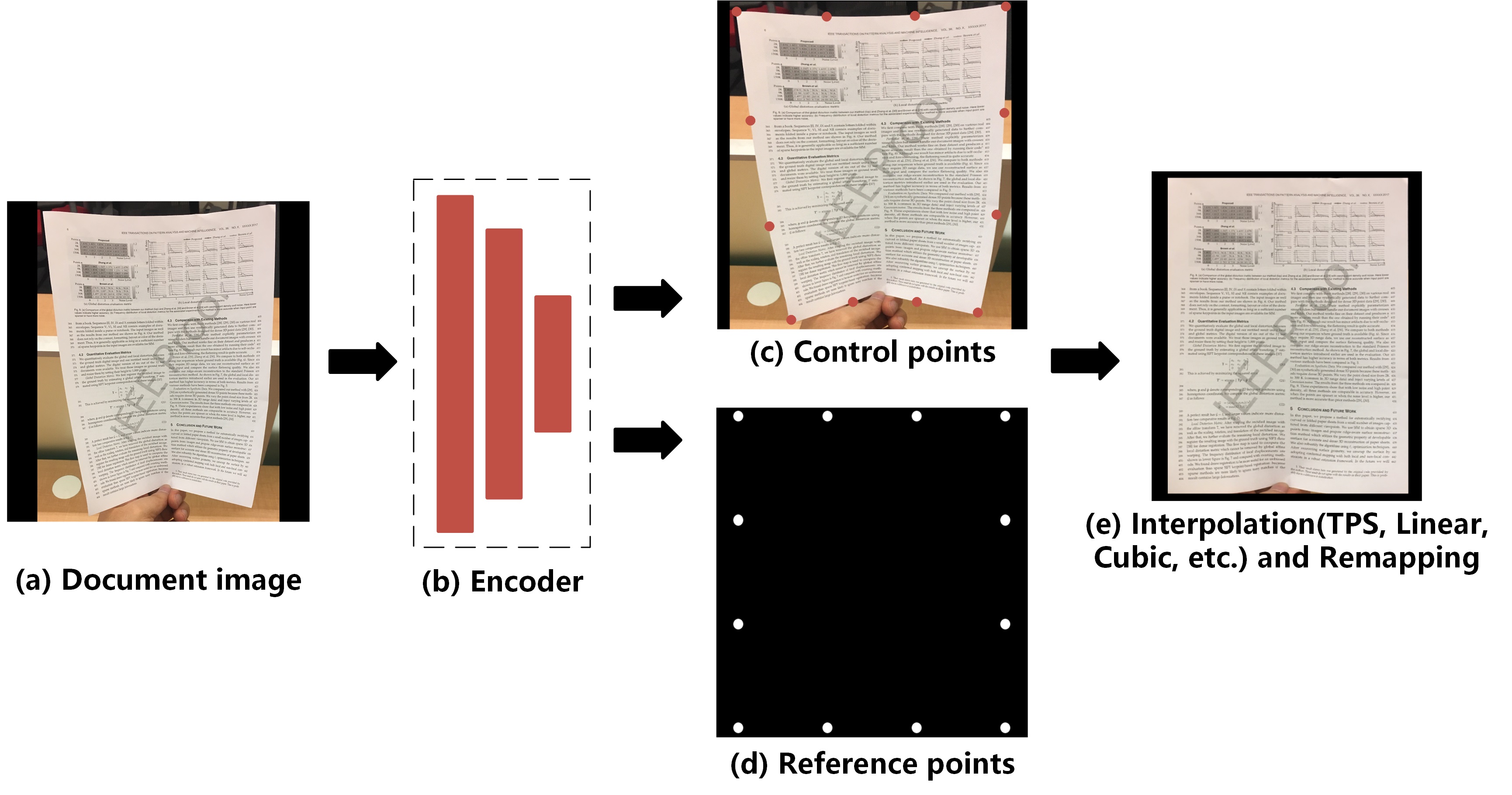}
\caption{{\bfseries Dewarping pipeline.} (a) input deformed document image. (b) Encoder architecture for extracting semantic information which  will be exploited to predict (c) control points and (d) reference points. Then, we convert sparse mappings to dense mapping and get the rectified image by (e) interpolation method and remapping respectively. Only 12 control points are used in this pipeline.} \label{fig2}
\end{figure}

\subsection{Definition}
Previous studies treated the geometric rectification task as a dense grid prediction problem, which take a 2D image as input and output a forward mappings  (each grid represents the coordinates of the pixels in the dewarped ouput image, and the pixels correspond to pixels in the warped input image) or backward mappings (each grid represents the coordinates of the pixels in the warped input image). Our method simplifies this process to directly predict the sparse mappings, and then uses interpolation to convert it into a dense backward mapping. To facilitate explanation, we define the following concepts:

{\bfseries Vertex} represents the coordinates of a point in document image. In this paper, we can move the vertices by changing the coordinates.

{\bfseries Control Points} consist of a set of vertices. As shown in Fig.~\ref{fig2} (c), control points are distributed on the distorted image to describe the geometric deformation of the document.

{\bfseries Reference Points} consists of the same number of vertices as the control points. As shown in Fig.~\ref{fig2} (d), the reference point describes the regular shape. Document Dewarping could be realized using unwarping grid by matching the control points and the reference points.

\subsection{Dewarping Process}

Fig.~\ref{fig2} illustrates the process in our work. First, an image of a deformed document is fed into network to obtain two output branches. Our approach adopts the Encoder architecture as feature extractor, and then exploits the learned feature to predict control points and reference points respectively in a multi-task manner. Second, as shown in Fig.~\ref{fig1}(e), we construct the rectified grid by moving the control points to the position of the reference points and converting it to pixel-wise location mapping. In order to move the position of the control points and convert sparse mappings to dense mapping, we employ interpolation method~\cite{other_5} (TPS, Linear, Cubic, etc.) between control points and reference points. After that, pixels are extracted from one place in the original distorted document image and mapped to another position of the rectified image. Compared with previous methods based on DNNs, our approach is simple and easy to implement.

\subsection{Network Architecture}

\begin{figure}
\includegraphics[width=\textwidth]{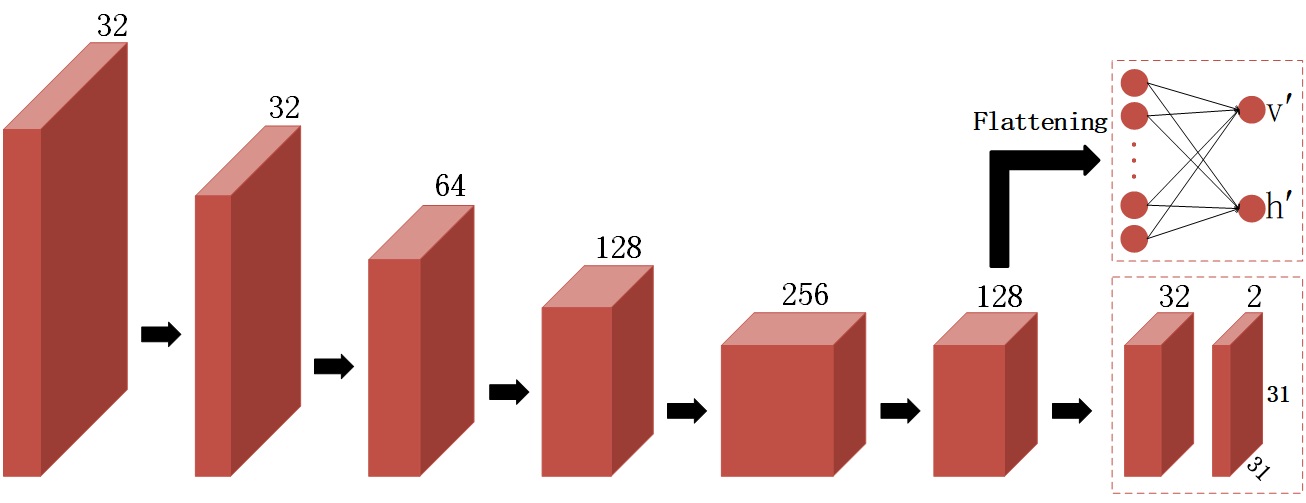}
\caption{{\bfseries Network architecture.} An encoder extracts image features and sends them to two branches. The upper branch is a fully connected neural network which predicts the interval between reference points. The lower branch is a two-layer convolutional network to predict control points.} \label{fig3}
\end{figure}

As shown in Fig.~\ref{fig2} and Fig.~\ref{fig3}, our approach takes the image of deformed document as an input and predicts control points $\mathbb{R} ^{31\times 31\times 2}$ and reference points $\mathbb{R} ^{31\times 31\times 2}$. Control points consist of ${31\times 31}$ coordinates to match the same number of reference points so as to construct the rectified grid. Since the reference points are composed of a regular grid, they can be constructed by the intervals of points between the horizontal and vertical directions $\binom{v^{'} }{h^{'} }$.

In our network, the first two layers of encoder use two convolutional layers with the strides of 2 and 3x3 kernels. Inspired by the architecture from~\cite{ref_16}, we use the same structure in our encoder architecture, including the dilated residual block and the spatial pyramid with stacked dilated convolution. After each convolution, the Batch Normalization and ReLU are applied. Then, we flatten the features of the last layer and feed them into the fully connected network to predict the intervals $\binom{v^{'} }{h^{'} }$ between reference points. Simultaneously, we use two-layer convolutional network to predict control points $\mathbb{R} ^{31\times 31\times 2}$, which apply Batch Normalization and PReLU after the first convolution.

\subsection{Training Loss Functions} 
We train our models in a supervised manner by using synthetic reference points and control points as the ground-truth. Training loss functions are composed of two parts. One is used to regress the position of the point, and the other is the interval between two points in the horizontal and vertical directions.

The Smooth L1 loss~\cite{other_6,other_7} is used for position regression on control points, which is less sensitive to outliers. It is defined as:

\begin{equation}
z_{i} = \begin{cases}
 0.5\left ( p_{i}- \hat{p_i}\right)^{2} , & \text{ if } \left |  p_{i}- \hat{p_i} \right |<1 \\
 \left |  p_{i}- \hat{p_i} \right |-0.5 ,  & \text{ otherwise }
\end{cases}
\end{equation}

\begin{equation}
L_{smoothL1}=\frac{1}{N_c} \sum_{i}^{N_c} z_{i} ,
\end{equation}
${N_c}$ is the number of control points, $p_i$ and $\hat{p_i}$ respectively denote the ground-truth and predicted the position in 2D grid.

\begin{figure}
\includegraphics[width=\textwidth]{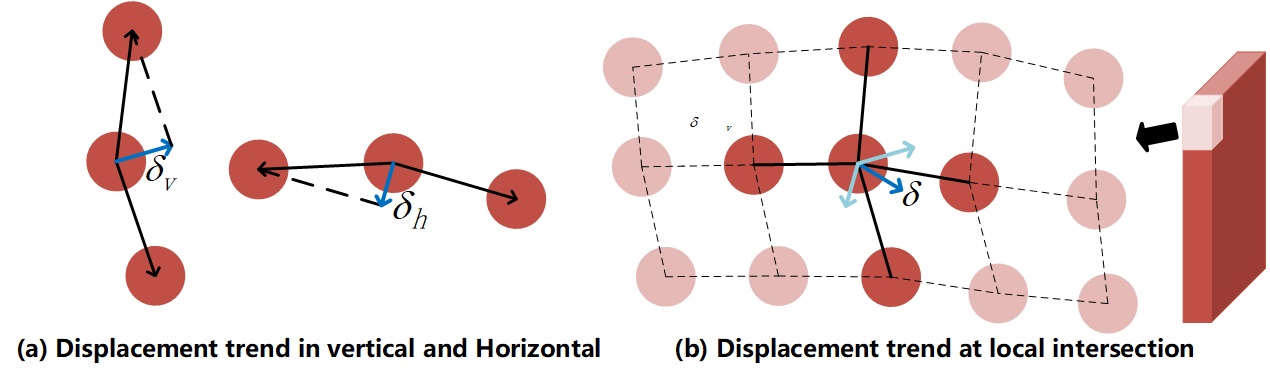}
\caption{{\bfseries Differential Coordinates.} We use differential coordinates as an alternative representation of vertex coordinates. $\delta _{v} $ and $\delta _{h}$ respectively represent the displacement trend of the center point in the horizontal and vertical.  $\delta $ represents the local positional correlation between the intersection and its connection points} \label{fig4}
\end{figure}

Different from facial keypoints detection, the content layout of document images is irregular. Although the Smooth L1 loss guide the model on how to place each vertex in an approximate position, it is difficult to represent the relationship of a point relative to its neighbors or local detail at each surface point. To make a more fault-tolerant model to better describe the shape, we use differential coordinates as an alternative representation for the center coordinates. As shown in Fig.~\ref{fig4}, $\delta _{v} $ and $\delta _{h}$ respectively represent the displacement trend of the center point in the horizontal and vertical, which helps to maintain correlation in the corresponding direction. Similarly, $\delta $ represents the local positional correlation between the intersection and two directions, which can be defined as :
\begin{equation}
\delta = \sum_{j=1}^{k}( p_j - p_{i}) ,
\end{equation}
$k$ is the number of elements, $p_{i}$ is the intersection. Inspired by~\cite{ref_16}, we use a constraint to expect the predicted displacement trend at local intersection to be as close as possible to the ground-truth. The displacement trend represents the relative relationship between a local region and its central point. We formulate $L_{c}$ function as follows:
\begin{equation}
\begin{aligned}
L_{c} &= \frac{1}{N_c}\sum_{i}^{N_c} \left ( \delta_i - \hat{\delta_i}\right )^{2}
\end{aligned}
\end{equation}

\begin{figure}
\includegraphics[width=\textwidth]{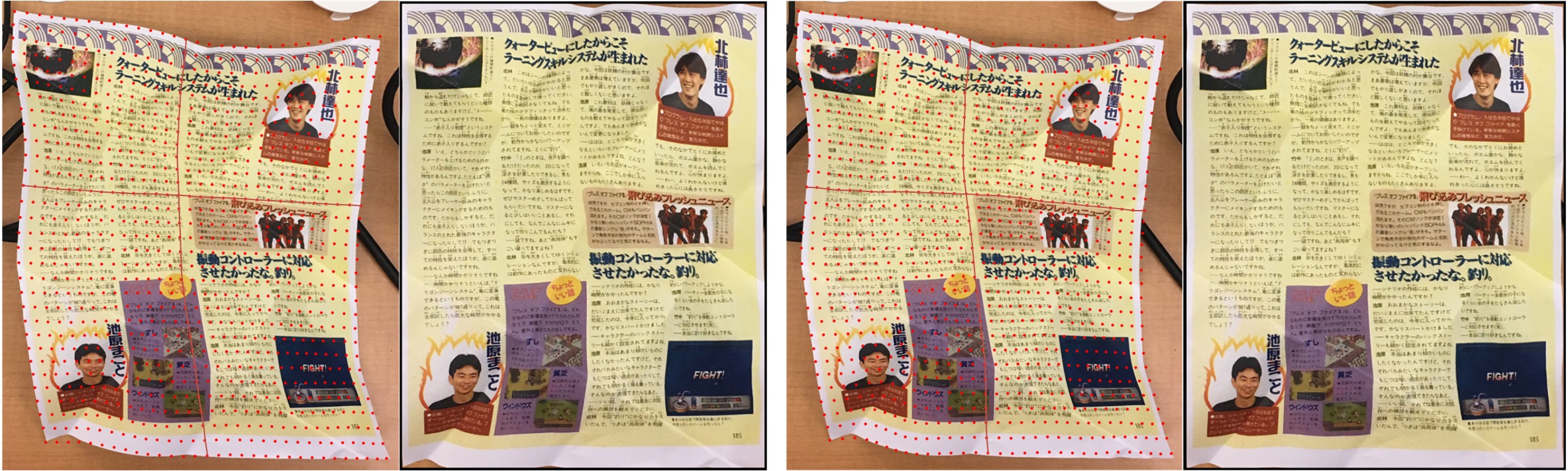}
\caption{{\bfseries Wide-Area Correlation.} To make the model better describe the shape, we adjust the number of constrained related vertices. The first group associates 4 neighbor vertices and sets k for 5. The second group is the center point connected directly with 4 vertices in each direction and sets k for 17. } \label{fig8}
\end{figure}

The predictions of control points are not independent of each other by using correlation constraints between vertices. In Fig.~\ref{fig4} (b), we associate 4 directly connected vertices and set $k=5$, which constrains the relationship between a point and its neighbors. We can constrain vertices in wide-area by expanding the correlation points in all four directions simultaneously. As shown in  Fig.~\ref{fig8}, we associate four vertices in each direction, which allows the model better learn the shape of the document image.

We use L1 loss to perform interval regression on reference points, which is defined as: 
\begin{equation}
L_r=\frac{1}{N_r} \sum_{i}^{N_r} \| d_i -  \hat{d_i}\|_1 ,
\end{equation}
where $N_r = 2$, $d$ represents the interval between two points in the horizontal or vertical directions. The final loss is defined as a linear combination of different losses:
\begin{equation}
L=L_{smoothL1} + \alpha L_{c} + \beta L_{r} ,
\end{equation}
where $\alpha$ and $\beta$ are weights associated to $L_{c}$ and $L_{r}$.

\subsection{Interactivity}

Control points are flexible and controllable, which facilitates the interaction with people to change the resolution of the rectified image, choose the number of vertices and adjust the sub-optimal vertices. Furthermore, our method can also be used to produce annotations for distorted document image.

{\bfseries Resize the resolution of the rectified image.} The smaller input image requires less computing, but some information maybe lost or unreadable. For facilitating implementation, we use smaller image as input to get control points and adjust these as follows:
\begin{equation}
O_{new}=F\left (\mathcal{M} / R_{old} * R_{new};I_{new} \right ) ,
\end{equation}
where $\mathcal{M} \in \mathbb{R}^{31\times 31\times 2}$ is original control points, $R_{old}$ and $R_{new}$ respectively denote original image resolution and adjusted resolution, $F$ is the interpolation and remapping, $I_{new}$ and $O_{new}$ are the distorted image and the rectified image with new-resolution. In this way, we can freely change resolution of the rectified image.

\begin{figure}
\includegraphics[width=\textwidth]{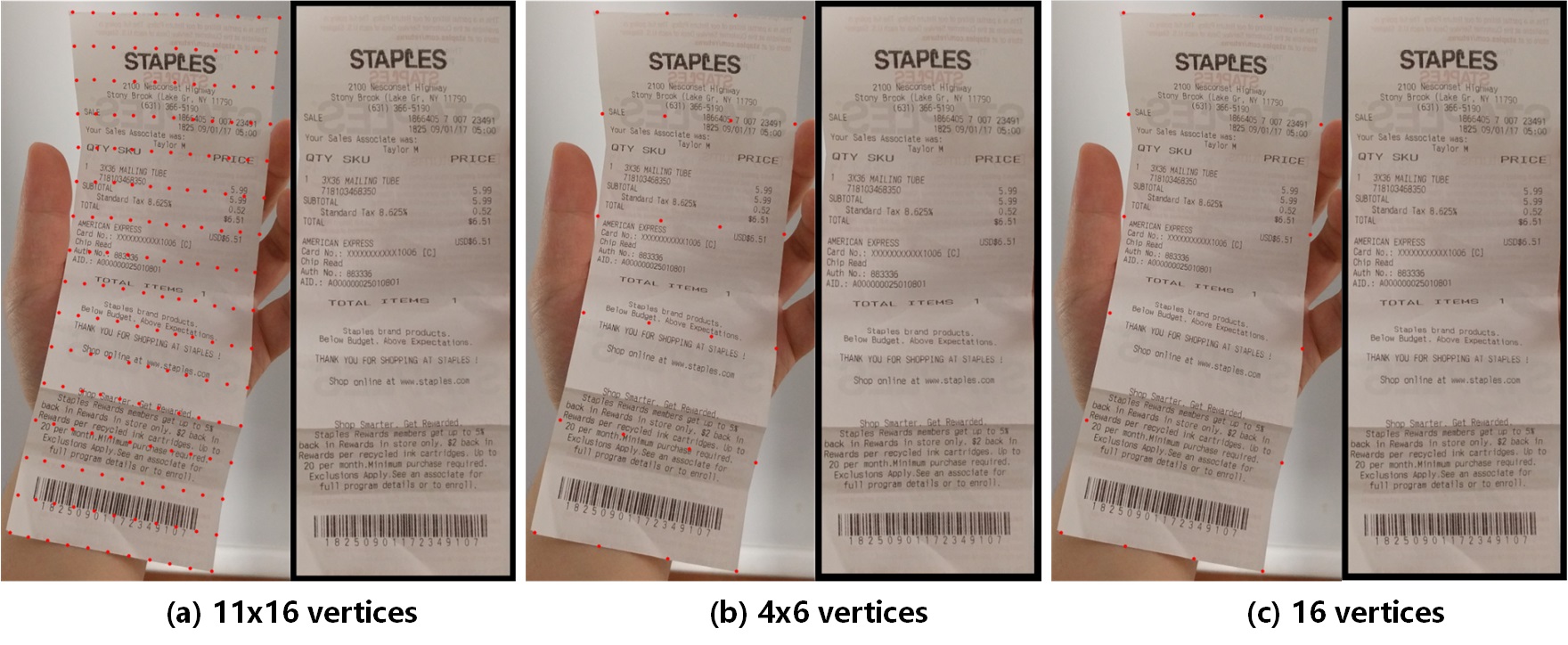}
\caption{{\bfseries Sparse vertices.} We can choose the number of vertices according to step length in Table.~\ref{tab1}. For simple deformations, a small number of control points can be used to achieve similar results. } \label{fig5}
\end{figure}

{\bfseries Choose the number of vertices.} Control points consist of ${31\times 31}$ coordinates, which can rectify a variety of distorted images. When using TPS interpolation in Fig.~\ref{fig2} (e), more vertices can take more deformation information into account, but more calculation time is needed. As shown in Fig.~\ref{fig5}, we can get similar results with fewer control points. For datasets of different difficulty, we can select the appropriate number of vertices according to step length in Table.~\ref{tab1}. When we rectify simpler distorted document image in Fig.~\ref{fig2}, internal vertices can also be omitted and the dewarping could be realized quickly with only 12 control points.

\begin{table}
\caption{{\bfseries Select vertices by step length.} The number of {\bfseries vertices} corresponding to the {\bfseries step} length. When we use a small number of vertices, the speed of calculation can be further improved in the post-processing steps.}\label{tab1}
\centering
\begin{tabular}{M{2cm}|M{0.6cm}M{0.6cm}M{0.6cm}M{0.6cm}M{0.6cm}M{0.6cm}M{0.6cm}M{0.6cm}}
\hline\noalign{\smallskip}
{\bfseries step} &  1 & 2 & 3  & 5 & 6 & 10 & 15 & 30 \\
\noalign{\smallskip}
\hline
\noalign{\smallskip}
{\bfseries vertices} & 31 & 16 & 11 & 7 & 6 & 4 & 3 & 2 \\
\noalign{\smallskip}\hline
\end{tabular}
\end{table}

\begin{figure}
\includegraphics[width=\textwidth]{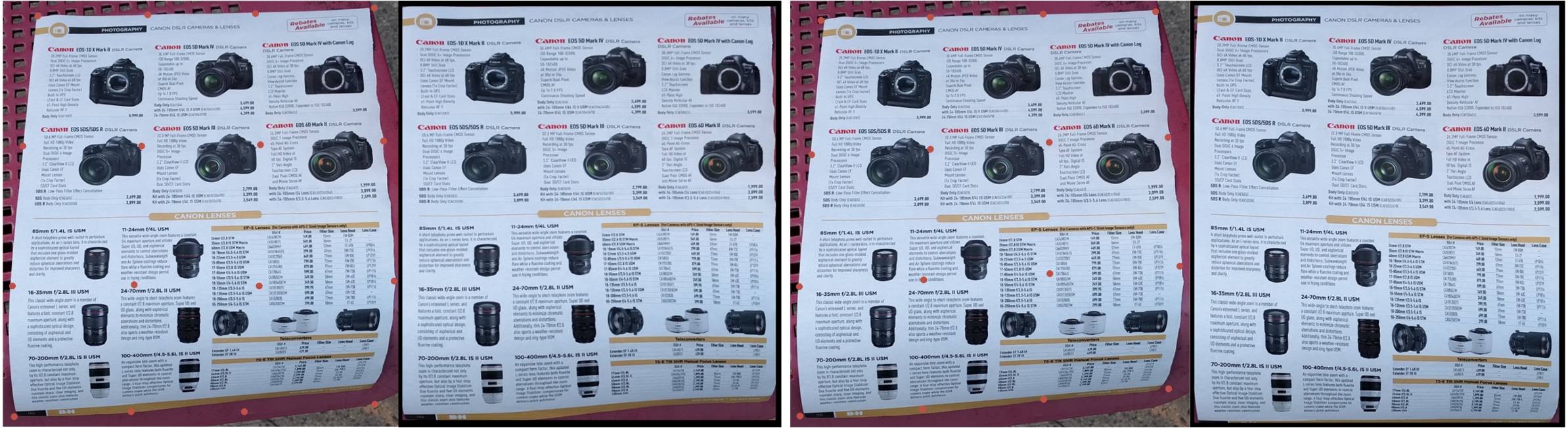}
\caption{{\bfseries Move the vertices.} The first group is the predicted initial position ($4 \times 4$ vertices). The second group is the adjusted position. By adjusting the coordinates of the sub-optimal vertices, we can get a better rectified image.} \label{fig6}
\end{figure}

{\bfseries Adjust the sub-optimal vertices.} Although generating more realistic training dataset can improve generalization in real-world images, it is difficult to ensure that all images can be well rectified with limited data. In addition to exploring more excellent and robust models, manual fine-tuning of the sub-optimal vertices is a more intuitive method. As shown in Fig.~\ref{fig6}, we drag control point to new locations or edit Laplacian mesh~\cite{other_1} to change the mesh's shape. Like~\cite{other_2}, a similar approach can also be used to automatically adjust the control points by iteratively optimizing the shallow neural network.

{\bfseries Distorted document annotation via control points.} Our method can be used as a semi-automatic annotation tool for distorted document image. Regardless of whether the rectifying systems shown in Fig.~\ref{fig1} use forward mapping, backward mapping or control points as the middleware to rectify image, these methods face the problem of difficulty in annotation. Existing methods and recent advances focuse on generating more realistic training dataset to improve generalization in real-world images, but it is difficult to rely on rendering engines to simulate the real environment. Our method provides a feasible way for distorted document annotation. Due to the flexibility and controllability of control points, it is convenient to interact with people and adjust the sub-optimal points. Satisfactory control points can also get forward mapping and backward mapping by interpolation.

\section{Experiments}

We synthesize 30K distorted document image for training. The network is trained with Adam optimizer and hyperparameters are set as $\alpha=0.1$ and $\beta=0.01$. We set the batch size as 32 and learning rate as $2e-4$ which was reduced by a factor of 0.5 after each 40 epochs.

\subsection{Datasets}
\begin{figure}
\includegraphics[width=\textwidth]{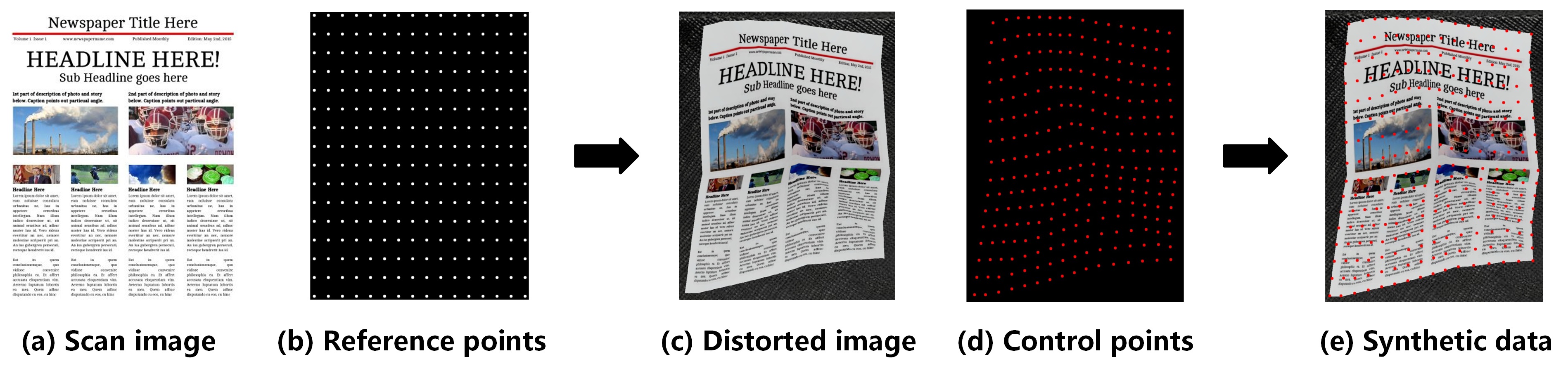}
\caption{{\bfseries Synthesize distorted document image.} We uniformly sample a set of (b) reference points on (a) scanned document image, and then perform geometric deformation on them to get (c) distorted image and (d) control points. (e) Synthetic data consists of distorted image, reference points and control points. } \label{fig9}
\end{figure}

Inspired by~\cite{ref_11,ref_16}, we synthesize 30K distorted document image in 2D mesh. The scanned document such as receipts, papers, etc using two functions proposed by~\cite{ref_8} to change the distortion type, such as folds and curves. We augment the synthetic images by adding random shadows, affine transformation, gaussian blur, background textures, jitter in the HSV color space and resize them into 992x992 (keeping the aspect ratio and zooming in or out along the longest side, then filling zero for padding). Meanwhile, we perform the same geometric transformation on a sparse grid $\mathbb{R} ^{61\times 61\times 2}$ to get the control points. As shown in Fig.~\ref{fig9}, a sparse grid is a set of points with intervals and evenly distributed on the scanned document. In this way, the ground-truth of control points and reference points can be obtained respectively. We can directly predict the coordinates of the control points. Similarly, we can also convert them into the position offset of the corresponding point, which is used to supervise the offset of the control point and then convert it into coordinates. In the experiment, these two groups of supervision methods have similar effects. In addition, the number of vertices in control points can be selected according to requirements. As shown in Table.~\ref{tab2}, we can choose different step length to change the density of the grid and the length of the interval. In our work, the step size is set to 2.

\begin{table}
\caption{Select the sparseness of vertices in the grid through different step length.}\label{tab2}
\centering
\begin{tabular}{M{2cm}|M{0.6cm}M{0.6cm}M{0.6cm}M{0.6cm}M{0.6cm}M{0.6cm}M{0.6cm}M{0.6cm}M{0.6cm}M{0.6cm}M{0.6cm}M{0.6cm}}
\hline\noalign{\smallskip}
{\bfseries step} &  1 & 2 & 3 & 4 & 5 & 6 & 10 & 12 & 15 & 20 & 30 & 60 \\
\noalign{\smallskip}
\hline
\noalign{\smallskip}
{\bfseries vertices} &  61 & 31 & 21 & 16 & 13 & 11 & 7 & 6 & 5 & 4 & 3 & 2 \\
\noalign{\smallskip}\hline
\end{tabular}
\end{table}

\begin{figure}
\includegraphics[width=\textwidth]{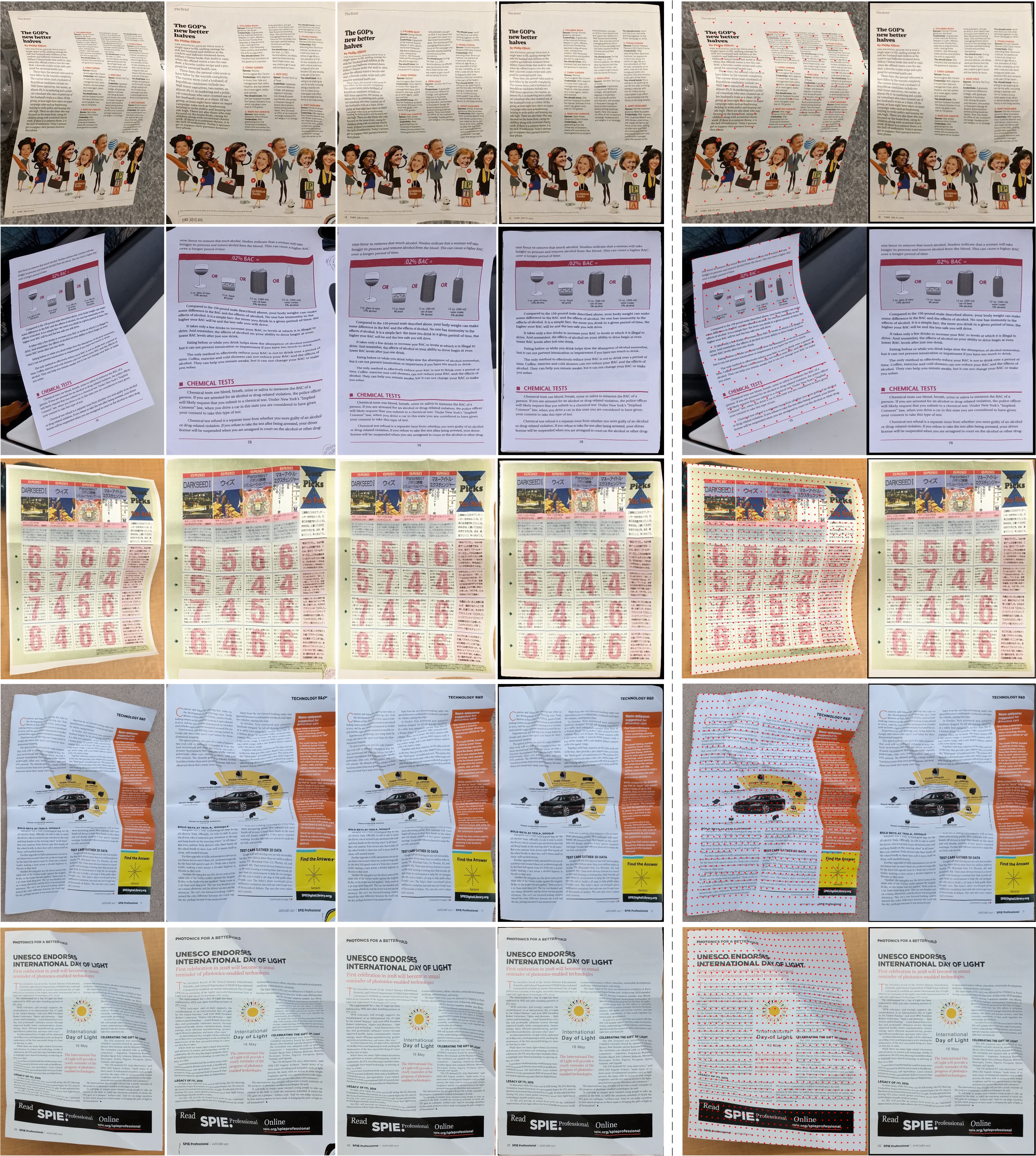}
\caption{{\bfseries Results on the Ma et al.~\cite{ref_11} benchmark dataset.} Col 1 : Original distorted images, Col 2 : Results of Ma et al.~\cite{ref_11}, Col 3 : Results of Das and Ma et al.~\cite{ref_12}, Col 4 : Results of Xie et al.~\cite{ref_16}, Col 5 : Position of control points, Col 6 : Results of our method. In the first two Row, our method uses $11 \times 16$ vertices to rectify distortion. The last three Row use $31 \times 31$ vertices.} \label{fig7}
\end{figure}

\subsection{Results}
We train network on synthetic data and test it on the benchmark dataset used by Ma et al.~\cite{ref_11} which has various real-world distorted document images. Compared with previous method, control points rectify distortions while removing background. In addition, the final backward mapping obtained by interpolation from sparse control points can effectively avoid local pixel jitter. As shown in Fig.~\ref{fig7}, the pixel-wise regression methods~\cite{ref_11,ref_12} pay attention to the global and local relations ,which often ignore the correction of document edge area. Although the multi-task method~\cite{ref_16} reduces edge blur by adding foreground segmentation tasks, the edge of most rectified image was still not neat enough. Our proposal addresses the difficulties in better balancing global and local. Control points focus on the global and interpolation method refines the local. These make our method simpler and more flexible while ensuring the correction effect.

\begin{table}
\caption{A comparison of different real-world distorted document images was made on the Ma et al.~\cite{ref_11} benchmark dataset.}\label{tab3}
\centering
\begin{tabular}{M{3cm}|M{3cm}M{3cm}}
\hline\noalign{\smallskip}
{\bfseries Method} &  {\bfseries MS-SSIM $\uparrow$} & {\bfseries LD $\downarrow$}\\
\noalign{\smallskip}
\hline
\noalign{\smallskip}
Ma et al.~\cite{ref_11} &  0.41 & 14.08 \\
Liu et al.~\cite{ref_14} &  0.4491 & 12.06 \\
Das and Ma et al.~\cite{ref_12} &  0.4692 & 8.98 \\
Xie et al.~\cite{ref_16} & 0.4361 & {\bfseries 8.50}\\
Our & {\bfseries 0.4769} &  9.03\\
\hline
\end{tabular}
\end{table}

We use Multi-Scale Structural Similarity (MS-SSIM)~\cite{other_3} and Local Distortion (LD)~\cite{other_4} for quantitative evaluation. MS-SSIM evaluates the global similarity between the rectified image and scanned image in multi-scale. LD computes the local metric by using SIFT flow between the rectified image and the ground truth image. The quantitative comparisons between MS-SSIM and LD are shown in Table.~\ref{tab3}. Our method demonstrates state-of-the-art performance in the quantitative metric of global similarity, and slightly weaker than the best method in local metric.

\begin{table}
\caption{Compare interpolation method and number of control points on the Ma et al.~\cite{ref_11} benchmark dataset.}\label{tab4}
\centering
\begin{tabular}{M{2cm}M{2cm}|M{3cm}M{2cm}M{2cm}}
\hline\noalign{\smallskip}
{\bfseries Interpolation} & {\bfseries Vertices} &  {\bfseries MS-SSIM $\uparrow$} & {\bfseries LD $\downarrow$} & {\bfseries Time}\\
\noalign{\smallskip}
\hline
\noalign{\smallskip}

TPS & $11 \times 16$ & {\bfseries 0.4769} &  {\bfseries 9.03} & 330ms\\
TPS & $4 \times 6$ &  0.4694 &  9.19 & 61ms\\
TPS & 16 &  0.4638 &  9.58 & 46ms\\

Linear & $31 \times 31$ &  0.4757 &  9.08 & {\bfseries 25ms}\\
Linear & $16 \times 16$ &  0.4757 &  9.09 & 25ms\\
\hline
\end{tabular}
\end{table}

To trade-off between computational complexity and rectification performance, we compare different interpolation method and the number of vertices as shown in Table.~\ref{tab4}. Thin plate splines (TPS) are a spline-based technique for data interpolation and smoothing. When we use TPS interpolation, dense control points can better rectify complex distortions, but it also increases the amount of calculation. In contrast, linear interpolation requires less calculation, but the effect is slightly worse. We run our network and post-processing on a NVIDIA TITAN X GPU which processes 2 input images per batch and Intel(R) Xeon(R) CPU E5-2650 v4 which rectifies distorted image by using Linear interpolation in multiprocessing, respectively. Our implementation takes around 0.025 seconds to process a $992\times 992$ image.

\section{Conclusion}
In this paper, we propose a novel approach using control points to rectifying distorted document image. Different from pixel-wise regression, our method exploits the encoder architecture to predict control points and reference points, so that the complexity of the neural network is reduced greatly. Control points facilitates the interaction with people to change the resolution of the rectified image, choose the number of vertices and adjusts the sub-optimal vertices, which are more controllable and practical. Furthermore, our method is more flexible in selecting post-processing methods and the number of vertices. The control points can also be used as a preprocessing step to realize semi-automatic distorted document image annotation with some further control point editing by humans. Although our approach has better tradeoff between computational complexity and rectification performance, further research is needed to explore more lightweight and efficient models.

\section*{Acknowledgements}
This work has been supported by the National Key Research and Development Program Grant 2020AAA0109702, the National Natural Science Foundation of China (NSFC) grants 61733007, 61721004.

%
%
%
%
\bibliographystyle{splncs04}
\bibliography{samplepaper}

\end{document}